%% file: main.tex
\documentclass[10pt,twocolumn,letterpaper]{article}

\usepackage{cvpr}              
\usepackage[accsupp]{axessibility} 
\usepackage{algorithm}
\usepackage{algpseudocode}
\definecolor{cvprblue}{rgb}{0.21,0.49,0.74}
\usepackage[pagebackref,breaklinks,colorlinks,allcolors=cvprblue]{hyperref}
\usepackage{tikz}
\usepackage{pgfplots}
\pgfplotsset{compat=1.18}
\def\paperID{*****} 
\def\confName{CVPR}
\def\confYear{2026}

\title{HarmoniDiff-RS: Training-Free Diffusion Harmonization for\\  Satellite Image Composition}

\author{
    Xiaoqi Zhuang \quad Jefersson A. Dos Santos\thanks{Corresponding author.} \\
    The University of Sheffield \\
    {\tt\small \{xzhuang5, j.santos\}@sheffield.ac.uk}
    \and
    Jungong Han \\
    Tsinghua University \\
    {\tt\small jghan@tsinghua.edu.cn}
}

\begin{document}
\maketitle

\input{sec/0_abstract}    
\input{sec/1_intro}

\input{sec/2_related_work}

\input{sec/3_method}

\input{sec/4_experiment}

\input{sec/5_ablation_study}

\input{sec/6_limitation}
\input{sec/7_conclusion}

\end{document}

%% file: sec/0_abstract.tex
\begin{abstract}
Satellite image composition plays a critical role in remote sensing applications such as data augmentation, disaste simulation, and urban planning.
We propose \textbf{HarmoniDiff-RS}, a training-free diffusion-based framework for harmonizing composite satellite images under diverse domain conditions.
Our method aligns the source and target domains through a \textbf{Latent Mean Shift} operation that transfers radiometric characteristics between them. To balance harmonization and content preservation, we introduce a \textbf{Timestep-wise Latent Fusion} strategy by leveraging early inverted latents for high harmonization and late latents for semantic consistency to generate a set of composite candidates. A lightweight harmony classifier is trained to further automatically select the most coherent result among them.
We also construct \textbf{RSIC-H}, a benchmark dataset for satellite image harmonization derived from fMoW, providing 500 paired composition samples. Experiments demonstrate that our method effectively performs satellite image composition, showing strong potential for scalable remote-sensing synthesis and simulation tasks. Code is available at: \url{https://github.com/XiaoqiZhuang/HarmoniDiff-RS}
\end{abstract}

%% file: sec/1_intro.tex
\section{Introduction}
\label{sec:intro}

\begin{figure}[t!]
    \centering
    \includegraphics[width=\linewidth]{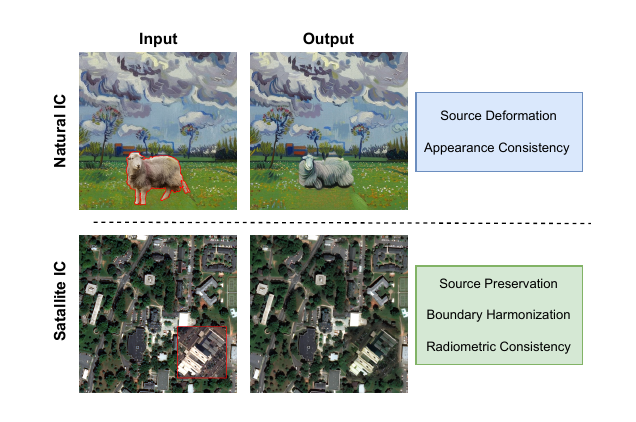}
    \caption{Comparison between natural~\cite{pham2024tale} (upper) and satellite (bottom) image composition.  Natural image composition emphasizes semantic alignment and may involve non-rigid source deformation, while satellite composition focuses on boundary harmonization under geometric rigidity.}
    \label{fig:SIC}
\end{figure}

Composing satellite images is a fundamental but under-explored problem in remote sensing. A reliable image composition framework could enable data augmentation, disaster simulation, and urban planning by synthesizing reasonable satellite scenes under various temporal, geographic, and environmental conditions. 
In natural image composition~\cite{Perez03Poisson, FreeCompose, pham2024tale, lu2023tf}, the main challenge lies in achieving semantic alignment between the well-segmented source foreground and the target background.To achieve visual realism, the source object is often allowed to undergo non-rigid geometric deformation or appearance adjustments, such as changing the pose or shape to better fit the surrounding context (e.g., the sheep in Fig.~\ref{fig:SIC} adapts from a standing to a sitting posture).
In contrast, satellite image composition focuses on structurally rigid sources such as buildings, roads, or ports, where geometric integrity must be preserved for the result to remain realistic.
The primary challenge therefore shifts from semantic alignment to boundary harmonization and radiometric consistency, ensuring seamless integration without distorting the original structure, as shown in Fig.~\ref{fig:SIC}. 
Therefore, in this paper, we define \textit{satellite image composition as the task of harmonizing a rigid source region with a target scene, ensuring boundary smoothness and appearance consistency without altering the original geometry}. Note that our definition of satellite image composition differs from its conventional use in remote sensing, where the term typically refers to multi-sensor or multi-temporal image fusion for radiometric normalization or mosaicking.

Despite significant progress in natural image harmonization and blending~\cite{Perez03Poisson, FreeCompose, pham2024tale, lu2023tf}, these methods are not directly applicable to satellite imagery.
First, they often rely on deforming the source geometry to better blend with the target, an operation unsuitable for the structurally rigid targets commonly found in satellite scenes such as buildings and roads.
Second, unlike natural image datasets with object-level segmentation masks, satellite imagery rarely provides accurate source delineations, making it challenging to define precise compositional boundaries.
These differences motivate the development of a harmonization-based framework that can handle rigid, structure-preserving compositions without requiring segmentation-level supervision.

To address the above challenges, we propose HarmoniDiff-RS, a training-free diffusion harmonization framework for satellite image composition.
Unlike existing approaches that rely on online optimization~\cite{FreeCompose, pham2024tale}, or fine-tuning~\cite{zhang2020deep, tan2023deep, yang2022paint}, our method operates entirely in the diffusion latent space to exploit generative priors for harmonization.
We first perform Latent Mean Shift Normalization, aligning the radiometric and stylistic statistics of the source latent with those of the target latent to ensure consistent illumination and color/spectral balance.
Then, inspired by the properties of the Denoising Diffusion Implicit Model (DDIM)~\cite{song2021denoising} inversion, we observe that early inverted latents yield more harmonious but less identity-preserving results, whereas later latents retain structure but suffer from boundary artifacts.
To balance these complementary effects, we propose a Timestep-wise Latent Fusion strategy that selectively combines early and late latents under an edge-aware mask, achieving both boundary smoothness and geometric preservation.
Finally, a lightweight Harmony Classifier evaluates each timestep’s output and selects the most harmonious composition as the final result.
Together, these components enable geometry-preserving and radiometrically consistent composition without any model training or optimization.

In summary, our main contributions are as follows:
\begin{itemize}
    \item \textbf{We define and address the novel task of realistic satellite image composition}, which differs from natural image composition by focusing on boundary harmonization and geometric preservation rather than semantic alignment or object deformation.  
    We further propose a \textit{training-free diffusion harmonization framework, HarmoniDiff-RS}, to tackle this task.
    
    \item \textbf{We introduce a diffusion-latent harmonization mechanism} that jointly performs radiometric alignment and timestep-wise latent fusion.  
    This mechanism leverages the complementary properties of early and late inverted latents to achieve both structural fidelity and boundary smoothness, resulting in visually consistent compositions without fine-tuning.
    
    \item \textbf{We construct a benchmark RSIC-H} for satellite image composition, which includes a new dataset with diverse cross-domain source–target pairs and two quantitative metrics, Harmony Score (HS) and Boundary Gradient Difference (BGD), for assessing compositional realism and boundary smoothness.
\end{itemize}

%% file: sec/2_related_work.tex
\section{Related Work}
\label{sec:Related Work}

\subsection{Remote Sensing Diffusion Models}
Diffusion models~\cite{NEURIPS2020_4c5bcfec, song2021denoising} have emerged as powerful generative frameworks that gradually denoise Gaussian noise into realistic images through iterative refinement. DDIM~\cite{song2021denoising} provides a deterministic sampling and inversion formulation, enabling bidirectional mapping between image and latent domains without retraining.  
This property makes DDIM particularly appealing for tasks requiring real-image manipulation, as it allows controlled editing and harmonization while preserving structural content.

Recently, several works have adapted diffusion models to remote sensing imagery.  
Text2Earth~\cite{Text2Earth} and DiffusionSat~\cite{khanna2024diffusionsat} proposed text-to-satellite generation models, capable of synthesizing satellite images based on text descriptions, and the image quality is far superior to that generated using the traditional text-to-image model. 
Beyond text-driven generation, diffusion models have introduced other controllability by conditioning on bounding boxes~\cite{tang2024aerogen}, street maps~\cite{sastry2024geosynth}, or sketches~\cite{tang2024crs}, enabling fine-grained spatial generation.  More recently, CC-Diff++~\cite{CCDiff} focuses on the issue of regional inconsistency in controlled generation, proposing cross-region constraints to improve spatial coherence. Although it mitigates disharmony during conditioned synthesis, CC-Diff++~\cite{CCDiff} still generates new images from noise rather than adapting or harmonizing real satellite imagery.  
In contrast, our work tackles the problem of \textit{satellite image composition}, where the goal is not to generate an entire scene but to seamlessly integrate the source patch and the target scene with consistent appearance and radiometric properties.

\subsection{Image Composition}

Image composition aims to seamlessly integrate a foreground object into a background image while maintaining visual realism and structural coherence.  
Early methods approached this problem through pixel-domain blending techniques such as Poisson image editing~\cite{Perez03Poisson}, which focus on smoothing the gradient transitions at the source–target boundary.  
While effective for local illumination consistency, these methods rely purely on low-level cues and cannot handle boundary seams.

With the advent of deep learning, learning-based image harmonization methods~\cite{zhang2020deep, tan2023deep} introduced neural network architectures that learn to align with foreground and background.  
However, they typically assume access to accurate segmentation masks and deformable foregrounds, making them well suited for natural images but ill-suited for rigid structures commonly found in satellite imagery.

More recently, diffusion-based image composition approaches have emerged as powerful alternatives, leveraging generative priors for both structure preservation and semantic coherence.  
FreeCompose~\cite{FreeCompose} formulates harmonization as the guidance difference between conditional and unconditional diffusion processes, effectively locating and refining disharmonious regions.  
Tale~\cite{pham2024tale} achieves fine-grained foreground–background blending through attention modulation, while TF-ICON~\cite{lu2023tf} aligns global illumination and color balance by text-conditioned diffusion guidance.  
Although these methods significantly advance natural image composition, their underlying assumptions: availability of precise segmentation masks and allowance for non-rigid object deformation, are not valid for remote sensing scenes, where the goal is to maintain geometric rigidity and radiometric consistency rather than semantic realism.

In this work, we extend diffusion-based harmonization to the domain of satellite imagery.  
Unlike previous methods that deform the source image or rely on segmentation-based alignment, our approach focuses on boundary harmonization and radiometric alignment while preserving the geometric integrity of the scene, enabling realistic composition of existing satellite images.

%% file: sec/3_method.tex
\section{Proposed Method}

\begin{figure*}[t]
    \centering
    \includegraphics[width=\textwidth]{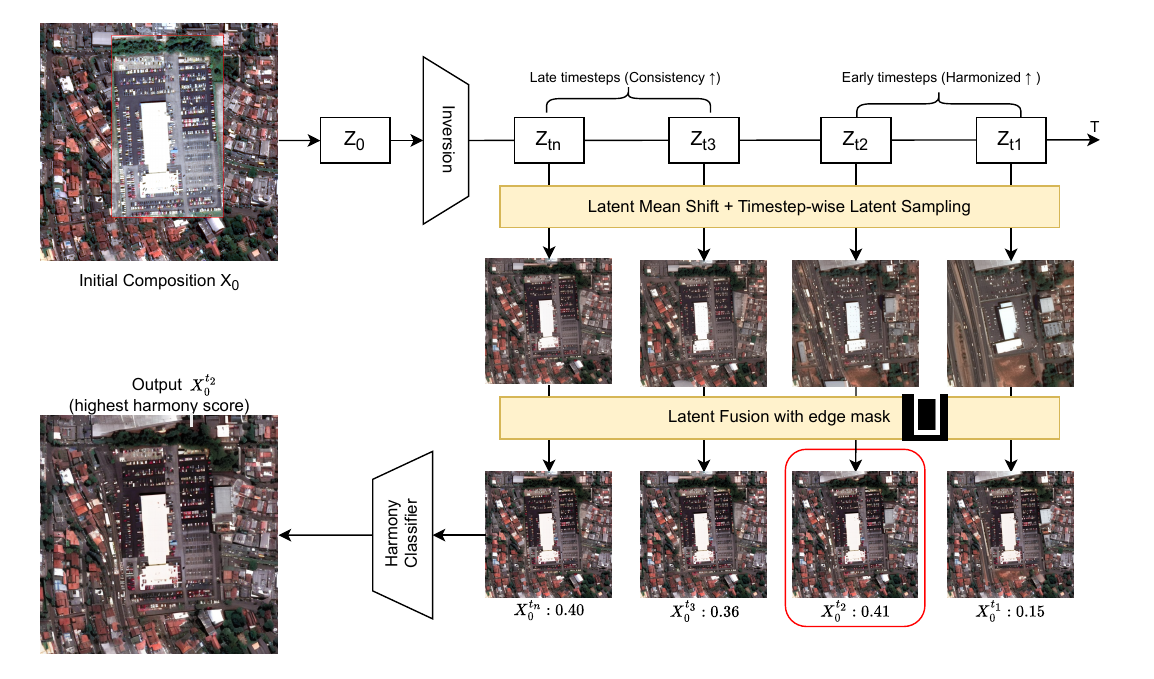}
    \caption{Overview of the proposed \textbf{HarmoniDiff-RS} framework. Given an initial composition of a target scene and a source patch, the model first applies \textit{Latent Mean Shift} to align the source latent with the target style. Then, \textit{Timestep-wise Latent Sampling} generates a sequence of intermediate compositions: early timesteps yield more harmonious but less faithful results, while late timesteps better preserve structural consistency but exhibit visible seams. Our \textit{Latent Fusion} combines both effects under an edge-aware mask, and the final output is selected based on the highest harmony score predicted by the harmony classifier.}
    \label{fig:Method}
\end{figure*}

\subsection{Preliminaries}

\paragraph{Latent Diffusion Models.} Our method builds upon latent-based diffusion models~\cite{rombach2021highresolution}, which operate in a latent space rather than pixel space by a pre-trained Variational Autoencoder~\cite{kingma2013auto} with an encoder $\mathcal{E}$ and a decoder $\mathcal{D}$. $z_0 = \mathcal{E}(x_0)$, $\qquad \hat{x}_0 = \mathcal{D}(z_0)$ where $x_0$ is the realistic image and $z_0$ is its representation in the latent space. 

\paragraph{DDIM Sampling and Inversion}
DDIM~\cite{song2021denoising, NEURIPS2020_4c5bcfec} sampling removes stochasticity during inference and then generates a deterministic latent trajectory $\{z_t\}^{T}_{t=0}$. DDIM sampling computes the denoised latent at timestep $t-1$ as:

\begin{equation}
\hat{z}_0(z_t, t) =
\frac{z_t - \sqrt{1-\bar{\alpha}_t}\,\epsilon_\theta(z_t, t)}
{\sqrt{\bar{\alpha}_t}},
\label{eq:ddim-scheduler}
\end{equation}
\begin{equation}
z_{t-1} =
\sqrt{\bar{\alpha}_{z-1}}\,\hat{z}_0(z_t, t)
+
\sqrt{1-\bar{\alpha}_{t-1}}\,\epsilon_\theta(z_t, t),
\label{eq:ddim-sampling}
\end{equation}
where $\epsilon_\theta(z_t,t)$ denotes the predicted noise from the diffusion model $\theta$, $\bar{\alpha}_t$ is the cumulative noise schedule.

To obtain the latent trajectory for a given real image $x_0$. We first get its representation $z_0$ in latent space by $z_0 = \mathcal{E}(x_0)$, and then DDIM inversion runs the sampling dynamics in reverse:
\begin{equation}
z_{t+1} =
\sqrt{\bar{\alpha}_{t+1}}\,\hat{z}_0(z_t, t)
+
\sqrt{1-\bar{\alpha}_{t+1}}\,\epsilon_\theta(z_t, t),
\label{eq:ddim-inversion}
\end{equation}
which recovers a latent trajectory $\{z_t\}_{t=0}^{T}$.

\subsection{Latent Mean Shift}

\begin{figure}[t]
    \centering
    \includegraphics[width=\linewidth]{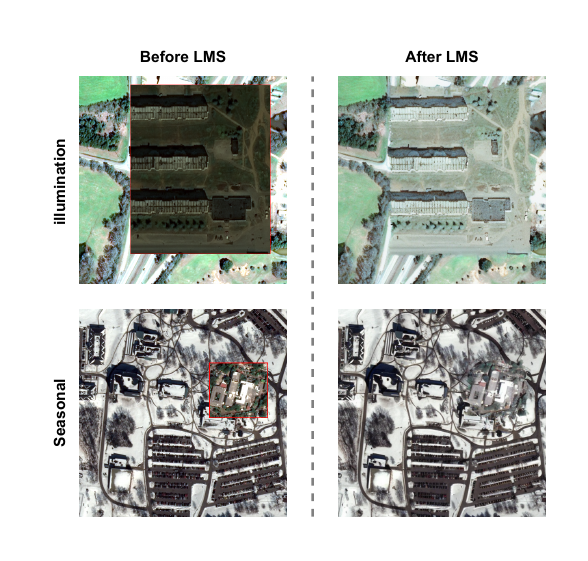}
    \caption{Effect of Latent Mean Shift (LMS) under different variations. Each column presents a composition example with distinct illumination or seasonal differences between the source patch and the target scene. The LMS module effectively aligns radiometric characteristics and color tones across acquisition conditions without retraining.
}
    \label{fig:LMS}
\end{figure}

Inspired by recent findings~\cite{pham2024tale, morita2024tkg} that latent channel statistics in diffusion models encode global appearance and style attributes, we hypothesize that channel-wise mean values serve as a lightweight style controller for satellite imagery. To harmonize source and target domains without any training, we apply a simple channel-wise mean alignment between the target latent and the source latent.
Given a source patch inversion latent $\text{src}_t \in \{\text{src}_t\}^T_{t=0}$, a target latent inversion latent $tar_t \in \{tar_t\}^T_{t=0}$, and a paste region $\Omega$, we compute per-channel means and shift the source latent such that its mean matches the target latent on the channel level:

\begin{align}
\Delta^c &= \mu_{\text{tar}_t^c} - \mu_{\text{src}_t^c} \\
\tilde{\text{src}}_t^c &= \text{src}_t^c + \Delta^c \\
\text{tar}_t^c(x,y) &=
\begin{cases}
\tilde{\text{src}}_t^c, & (x,y) \in \Omega \\
\text{tar}_t^c(x,y), & \text{otherwise}
\end{cases}
\label{eq:mean-shift}
\end{align}

Fig.~\ref{fig:LMS} demonstrates how Latent Mean Shift transfers the radiometric characteristics of the target image to the source patch. However, this step alone cannot remove hard boundaries, which we further address in the next sections.

\subsection{Timestep-wise Latent Fusion}

As illustrated in the upper part of Fig.~\ref{fig:Method}, DDIM-inverted latents at early timesteps (closer to noise) produce harmonious composites but suffer from weak condition preservation. In contrast, latents from late timesteps preserve the original structure and semantics of source and target images, yet result in noticeable hard boundaries. This observation motivates our design. Let $\text{tar}_t$ and $\text{src}_t$ denote the inverted latents of the target and source at timestep $t$. For each harmonious timestep $\text{ht} \in \{t_1, t_{2},...,t_n\}$, we apply the Latent Mean Shift to obtain the compositional latents $\{z_t^c\}^{tn}_{t1}$ by Eq.~\ref{eq:mean-shift}. We then perform DDIM sampling (Eq.~\ref{eq:ddim-sampling} for each latent until a pre-defined preserving timestep $t_{p}$, generating latent candidates $\{z_{p}^c\}^{tn}_{t1}$. Since inconsistencies mainly occur along hard edges after Latent Mean Shift, we construct an edge mask $M_{edge}$ around the source boundary, which is defined as:
\begin{equation}
M_{edge} = \text{dilate}(\Omega, w) - \text{erode}(\Omega, w)
\end{equation}
During the denoising process from $t_{p}$ to $t_0$, at each timestep we fuse harmonization latents $z_{t-1}^{edge}$ and identity-preserving $z_{t-1}^{p}$ as:
\[
z_{t-1}^{edge} = \text{DDIM}(z_t), z_{t-1}^{p} = \text{LatentMeanShift}(z_{t-1}^{src}, z_{t-1}^{tar})
\]
\begin{equation}
z_{t-1} = M_{edge} \cdot z_{t-1}^{edge} + (1-M_{edge}) \cdot z_{t-1}^{p}
\label{eq:zt}
\end{equation}
This process yields a set of final composite images $\{x_0^{t1}, x_0^{t1+1}, ..., x_0^{tn}\}$. 

\paragraph{Harmony Score Selection}
 However, different timesteps lead to different trade-offs between global harmonization and identity preservation, resulting that it is non-trivial to quantify and obtain the optimal $x_0$ from $\{x_0^{t1}, x_0^{t1+1}, ..., x_0^{tn}\}$. To automatically choose the best output, we train a lightweight remote-sensing harmony classifier $C_\psi$ that predicts whether a composite image is visually harmonious in the given region. The model takes as input a concatenation of the RGB image and a binary mask indicating the pasted region. Additional training details are provided in Sec.~\ref{sec:HC}. To increase the spatial robustness of the score, we evaluate the harmony score under three mask configurations: the original mask, a dilated mask, and an eroded mask. The final harmony score is the average of the three predictions. We then select the $x_0^{t}$ with the highest harmony score $s$ as our final composite output $x_0$

This pipeline allows our method to jointly leverage the harmonization ability of early inverted latents to fill boundary regions, while preserving semantic fidelity through late inverted latents, ensuring the final composition image remains faithful to the intended content. The complete Algorithm.~\ref{alg:method} of our method is visualized on Fig.~\ref{fig:Method}.

\begin{algorithm}[t]
\caption{Satellite Image Composition}
\label{alg:method}

\textbf{Input:} Target image $\text{TAR}$, source image $\text{SRC}$, source mask $\omega$, Harmony Classifier $C_\psi$, a set of harmonious timesteps $\text{ht} \in \{t_1, t_{2},...,t_n\}$, preservation step $t_{\text{p}}$ \\
\textbf{Output:} Final composite image $\hat{x}$

\begin{algorithmic}[1]

\State $w = \min(W_{\text{SRC}}, H_{\text{SRC}}) \cdot W_{\text{TAR}} \cdot 0.1$ \Comment{Edge Width} 
\State $M_{edge} \gets \text{dilate}(\Omega, w) - \text{erode}(\Omega, w)$ \Comment{Edge Mask}

\State $\text{tar}_0 = \mathcal{E}(\text{TAR}), \text{src}_0 = \mathcal{E}(\text{SRC})$

\For{$\text{ht} = t_1$ \textbf{to} $t_n$}
    \State $\text{tar}_{\text{ht}}\ \gets \text{DDIM Inversion}(\text{tar}_0)$
    \State $\text{src}_{\text{ht}}\ \gets \text{DDIM Inversion}(\text{src}_0)$
    \State $z_{\text{ht}} \gets \text{LatentMeanShift}(\text{src}_{ht}, \text{tar}_{ht})$ \Comment{Eq.~\ref{eq:mean-shift}}
    \For{$t = {\text{ht}}$ \textbf{to} $t_{\text{p}}$}
        \State $z_{t-1}^{ht} \gets \text{DDIM}(z_t^{ht})$
    \EndFor
    \For{$t = t_{\text{p}}$ \textbf{to} $t_{0}$}
        \State $z_{t-1}^{edge} \gets \text{DDIM}(z_t^{ht}) $
        \State $z_{t-1}^{p} \gets \text{LatentMeanShift}(\text{src}_{t-1}, \text{tar}_{t-1})$ \Comment{Eq.~\ref{eq:mean-shift}}
        \State $z_{t-1}^{\text{ht}} \gets M_{edge} \cdot z_{t-1}^{edge} + (1-M_{edge}) \cdot z_{t-1}^{p}$
    \State $x_0^{\text{ht}} = \mathcal{D}(z_0^{\text{ht}}) $
    \State $s_{\text{ht}} \gets C_{\psi}(x_0^{\text{ht}}, \omega)$ \Comment{Harmony Score Selection}
    \EndFor
\EndFor

\State $t^* \gets \arg\max_{\text{ht} \in \{t_1, \dots, t_n\}} s_{\text{ht}}$ 
\State \Return $x_0^{t^*}$

\end{algorithmic}
\end{algorithm}

%% file: sec/4_experiment.tex
\section{Experiments}

\subsection{Experiments Setup}
\paragraph{Data Preparation.} To benchmark remote sensing composition, we introduce RSIC-H, a dataset of realistic source–target pairs constructed from the Functional Map of the World (fMoW) dataset~\cite{fmow2018}. RSIC-H covers diverse land cover, season, illumination, and sensor variations. We randomly sample 413 target scene images and 381 source images from fMoW. The Ground Sampling Distance (GSD) metadata provided in fMoW is used to avoid source–target scale mismatch. Specifically, we rescale each source patch according to its GSD before insertion, ensuring that the object sizes remain physically consistent with the target scene. The resulting dataset covers realistic cross-domain composition settings with variations in geographic region, acquisition time, illumination, and sensor configuration, including differences in sun angles, off-nadir geometry, scan direction, and cloud coverage.

\paragraph{Implementation Details.} We adopt DiffusionSat~\cite{khanna2024diffusionsat} as our base generative model, a tailored generative model finetuned on SD 2.1 Model~\cite{rombach2021highresolution} with remote sensing data. Considering that timestep-wise diffusion sampling increases inference time, and that our compositional setting already benefits from strong visual priors rather than synthesizing content from scratch, we reduce the number of sampling steps to 20 and set the conditional guidance scale 3.5. The prompt for a composition task $\{\text{SRC}, \text{TAR}, \Omega\}$ is constructed on the source label and the target country as the template: "A satellite image of a [source label] in [target country]". We also feed the metadata of the target scene to help the model preserve global radiometric characteristics. We designate timestep 6-14 as harmonious timestep sets in Algorithm~\ref{alg:method} and the preservation step set is 15-20. These settings balance harmonization quality and runtime efficiency. All experiments run on a single NVIDIA A100. 

\paragraph{Harmony Classifier.}\label{sec:HC} Our harmony classifier is a lightweight ResNet-18~\cite{He_2016_CVPR} trained to detect visual harmonization in the pasted region. The classifier takes as input the concatenation of the RGB image and the binary mask, and outputs the probability that the masked region in the input image is visually harmonious. The training positive samples consist of real background images from RSIC-H, while the training negative samples are generated by random copy-paste compositions and Poisson blending composites. We construct 20,000 samples with a 1:1 positive/negative ratio and train for 5 epochs on one NVIDIA A100. The harmony classifier is introduced mainly to automate result selection and facilitate quantitative evaluation;
in practical use, the most harmonious result can be manually chosen from the generated candidates.

\subsection{Evaluation Metrics}
\label{sec:metrics}

We employ three complementary metrics to comprehensively evaluate the quality and harmony of the composed satellite images: 
Fréchet Inception Distance (FID)~\cite{heusel2017gans}, Harmony Score (HS), and Boundary Gradient Difference (BGD).

\paragraph{Fréchet Inception Distance (FID).}
Following common generative evaluation protocols, 
we adopt the FID metric to measure the distributional distance between the generated compositions and real satellite images from fMoW~\cite{fmow2018}.
Lower FID indicates that the generated images are closer to the real data distribution in terms of perceptual realism. 

\paragraph{Harmony Score (HS).}
To automatically assess visual consistency, we train a lightweight remote-sensing harmony classifier $C_\psi$ that predicts whether a composed image is visually harmonious within the pasted region. 
A higher HS indicates better overall visual harmony.

\paragraph{Boundary Gradient Difference (BGD).}
To quantitatively evaluate boundary smoothness, we further introduce the \textit{Boundary Gradient Difference (BGD)} metric, 
which measures the discontinuity of image gradients across the pasted boundary.
Let $I$ denote the grayscale version of the composed image, and $\Omega$ the binary mask of the pasted region. 
We define the inner and outer narrow boundary rings as:
\begin{equation}
\Gamma_{\text{in}} = \Omega \setminus \operatorname{erode}(\Omega, w), \quad
\Gamma_{\text{out}} = \operatorname{dilate}(\Omega, w) \setminus \Omega,
\end{equation}
where $w$ is the margin width in pixels. 
Let $G = \lVert \nabla I \rVert_2$ be the gradient magnitude map obtained via the Sobel operator.
The absolute version of BGD is defined as:
\begin{equation}
\text{BGD}_{\text{abs}} = 
\big|\mu_{\Gamma_{\text{in}}}(G) - \mu_{\Gamma_{\text{out}}}(G)\big|
\end{equation}
A smaller $\text{BGD}_{\text{abs}}$ indicates more consistent gradients between the inner and outer boundary regions, suggesting smoother and more harmonious transitions. In our experiments, we set $w=3$ pixels.

\subsection{Experiments Results}
\paragraph{Qualitative Results.}
\begin{figure*}[t]
    \centering
    \includegraphics[width=\textwidth]{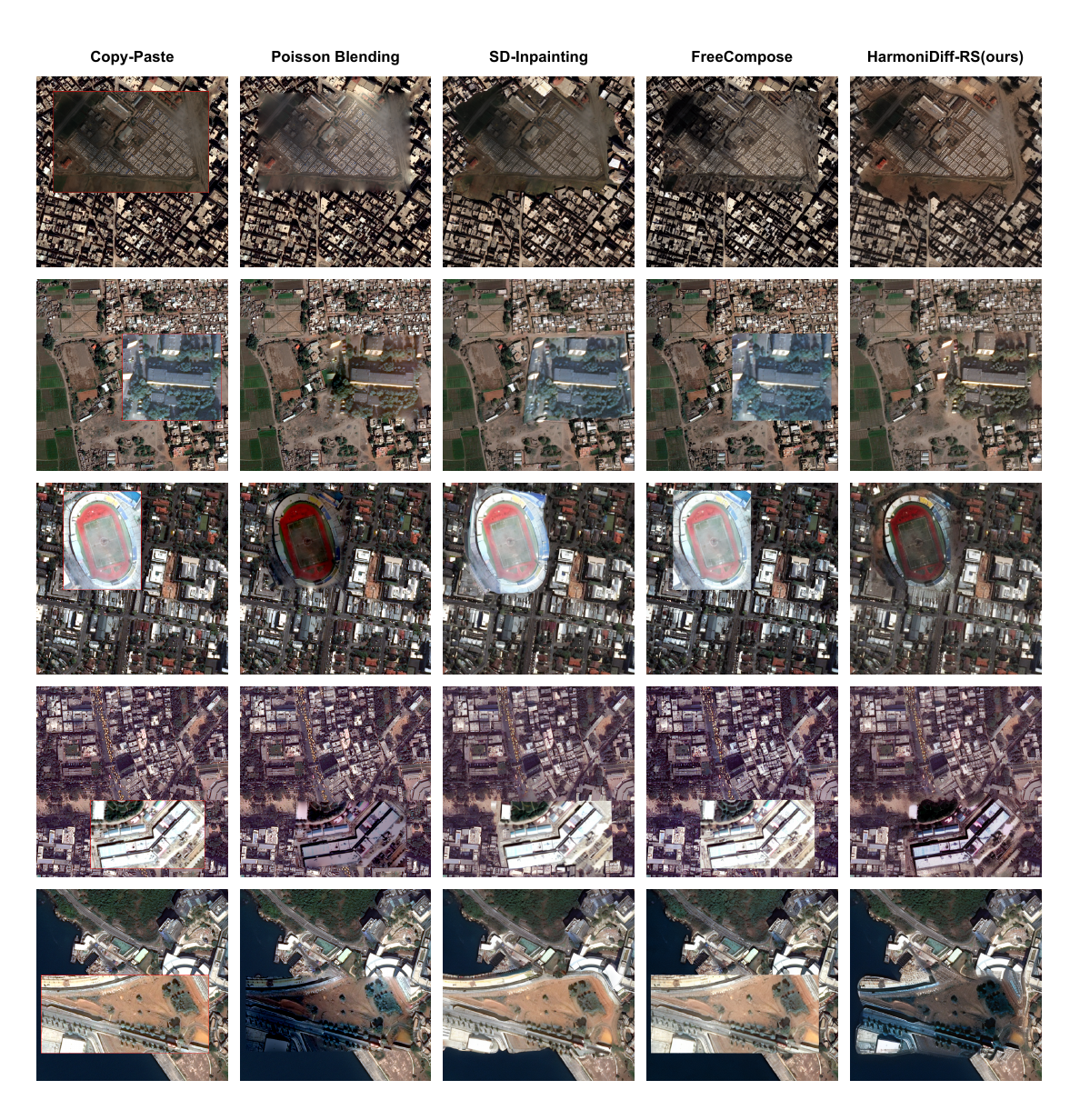}
    \caption{Qualitative comparison on RSIC-H. Our method generates seamless and semantically aligned compositions, outperforming copy-paste, Poisson blending, SD2 inpainting, and FreeCompose. Hard cases demonstrate our ability to correct semantic mismatches and produce realistic remote-sensing scenes.}
    \label{fig:Qualitative}
\end{figure*}

Fig.~\ref{fig:Qualitative} presents qualitative comparisons between our method and multiple baselines. The leftmost column shows the direct copy-paste results, which exhibit obvious boundary artifacts and strong source–target inconsistency. Poisson blending~\cite{Perez03Poisson} (second column) partially transfers radiometric appearance from the target scene to the pasted region, but still leaves visible seams and fails to resolve semantic inconsistencies.

We also compare against an SD2~\cite{rombach2021highresolution}-based inpainting model (third column), motivated by our use of edge masking. However, since inpainting models typically preserve the pasted content as a hard prior and only modify pixels inside the masked region, they struggle to adjust the semantic and radiometric attributes of the source image. 

The fourth column shows results from FreeCompose~\cite{FreeCompose}, a recent harmonization method designed for natural images. As our remote-sensing source images do not come with instance segmentation masks, we provide rectangular masks as input. Despite this, FreeCompose produces almost no substantive change to the pasted content, indicating its limited ability to handle remotely-sensed domain shifts.

In contrast, \textbf{HarmoniDiff-RS} (rightmost column) effectively transfers the target domain characteristics to the pasted region, while eliminating hard boundaries through latent-guided blending. Our model generates semantically and radiometrically coherent content at the transition zone, resulting in realistic and seamless composite satellite scenes. Notably, in the challenging case in row 5, where the original source patch and target scene are semantically mismatched, only our method successfully adapts the pasted region into a visually consistent port scene. These examples highlight that traditional inpainting and harmonization approaches are not sufficient for remote-sensing compositing, where content must adapt not only stylistically but also semantically to the surrounding geographical context.

\paragraph{Quantitative Results.}
\begin{table}[t]
\centering
\caption{Quantitative Results.}
\label{tab:Quantitative}
\begin{tabular}{l c c c}
\toprule
\textbf{Model} & \textbf{FID} $\downarrow$ & \textbf{HS} $\uparrow$ & \textbf{BGD} $\downarrow$\\
\midrule
Copy-Paste & 94.44 & 0.058 & 33.93\\
Poisson Blending & \textbf{90.66} & 0.173 & 8.70\\
Poisson Blending + Vae & 96.07 & 0.184 & 7.04\\
SD2 Inpainting & 97.92 & 0.084 & 9.18 \\
SD2 + FreeCompose & 97.25 & 0.076 & 30.28 \\
SD2 + HarmoniDiff-RS & \underline{95.32} & \underline{0.217} & \underline{6.48}\\
Sat + HarmoniDiff-RS \textbf{(ours)} & \underline{92.38} & \textbf{0.225} & \textbf{4.88}\\
\bottomrule
\end{tabular}
\end{table}

Table~\ref{tab:Quantitative} reports the quantitative comparison against baselines. Our method achieves the best harmony score (HS), indicating that the inserted source regions are perceptually the most coherent with their surrounding context. our method also achieves the lowest Boundary Gradient Difference (BGD) among all baselines, indicating that it produces the smoothest and most visually consistent boundaries between the source patch and the target scene.

In terms of FID~\cite{heusel2017gans}, our approach ranks second overall, slightly behind Poisson Blending. This is expected, as Poisson Blending does not modify the input pixels and therefore preserves original high-frequency details, which FID strongly favors.

To further clarify this effect, we additionally evaluate a PB+VAE variant, where input images are passed through the same VAE encoder-decoder used in our pipeline firstly and then applied Poisson Blending. This variant experiences a significant degradation in FID, comparable to ours, confirming that the FID gap is primarily caused by VAE compression in current diffusion backbones rather than harmonization quality. Fig.~\ref{fig:VAE_Recon} visualizes this limitation, showing that high-frequency remote-sensing textures (e.g., roofs, roads, grass patterns) tend to be partially smoothed by the VAE bottleneck.

Overall, while FID penalizes VAE-processed outputs, our method consistently delivers the most harmonious and semantically coherent compositions, as indicated by the highest HS and lowest BGD.

%% file: sec/5_ablation_study.tex
\section{Ablation Study}

\begin{table}[t]
\centering
\caption{Ablation study. LMS: Latent Mean Shift; TWR: Timestep-wise Reconstruction; LTF: Late Timestep Fusion. }
\label{tab:ablation_add}
\begin{tabular}{lccc}
\toprule
\textbf{Variant} & \textbf{FID} $\downarrow$ & \textbf{HS} $\uparrow$ & \textbf{CLIP} $\uparrow$ \\
\midrule
INIT (copy-paste) &  \,\,94.44 &  \,\,0.06  & \,\,1 \\
+ LMS &  \,\,91.95  &  \,\,0.15  & \,\,0.95 \\
+ TWR &  \,\,117.81 &  \,\,0.74  & \,\,0.85 \\
+ LTF (ours)&  \,\,92.38  &  \,\,0.22 & \,\,0.95 \\
\bottomrule
\end{tabular}
\end{table}

Table~\ref{tab:ablation_add} presents an ablation study to validate our designs and analyze their functions.
The Latent Mean Shift (LMS) improves the overall image fidelity by aligning the source patch’s latent statistics with the target latent, leading to a clear reduction in FID.
Adding Timestep-wise Latent Reconstruction (TWR), which independently reconstructs the image from each inverted timestep and selects the one with the highest harmony score, results in higher perceptual harmony but a noticeable drop in visual fidelity and CLIP image similarity, indicating partial loss of identity information. 
Finally, our Late Timestep Fusion (LTF) combines the strengths of both, balancing fidelity and harmony by blending early (harmonized) and late (identity-preserving) latents with an edge-aware mask.
Although this incurs a slight degradation in FID, it achieves the highest harmony score among all variants.
Additionally, Table~\ref{tab:Quantitative} compares our method on DiffusionSat~\cite{khanna2024diffusionsat} and SD2~\cite{rombach2021highresolution}, showing that even the SD2-based version of HarmoniDiff-RS already outperforms other SD2 baselines, while the specialized DiffusionSat backbone yields further improvement.

%% file: sec/6_limitation.tex
\section{Limitation}

\paragraph{VAE reconstruction constraint.}

\begin{figure}[t]
    \centering
    \includegraphics[width=\linewidth]{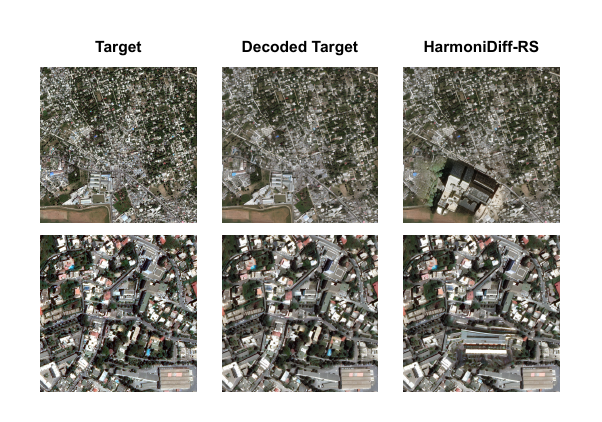}
    \caption{The target image (left) undergoes significant high-frequency loss after VAE encoding and decoding (middle left), which smooths out fine-grained textures such as roofs and roads. Our method (HarmoniDiff-RS, right) inherits this degradation because it operates in the latent space of the same VAE, resulting in slightly lower FID compared to pixel-domain methods like Poisson blending.}
    \label{fig:VAE_Recon}
\end{figure}

\begin{figure}[t]
    \centering
    \includegraphics[width=\linewidth]{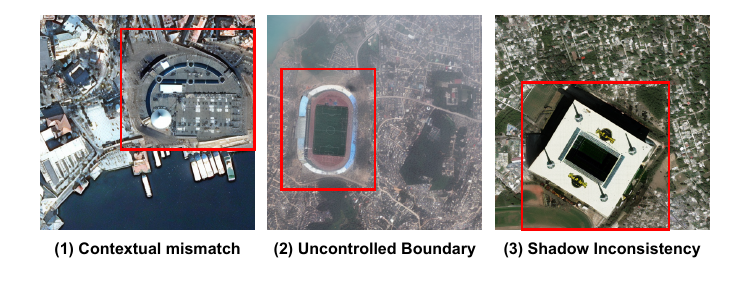}
    \caption{From left to right: (1) severe semantic and contextual mismatch between source patch and target scene; (2) lack of fine-grained control in boundary generation; and (3) shadow conditions that cannot be harmonized through our method.}
    \label{fig:Failure}
\end{figure}

Our approach operates in the latent space of diffusion models, and thus inherits the limitations of the underlying VAE. As shown in Fig.~\ref{fig:VAE_Recon}, remote-sensing textures with fine-grained structures (e.g., roof patterns, road lines, parking grids) are partially smoothed after VAE encoding and decoding. This compression reduces high-frequency fidelity and contributes to a moderate FID gap compared with non-generative baselines. We attribute this primarily to the reconstruction bottleneck of current diffusion backbones rather than our harmonization strategy. Importantly, our training-free pipeline can be directly plugged into future latent models with improved reconstruction capacity enabling immediate quality gains.
\paragraph{Failure case analysis.}
Failure cases are illustrated in Fig.~\ref{fig:Failure}. First, when the inserted source patch is severely semantically mismatched with the target scene, harmonization becomes challenging and may produce unnatural structures. Second, our current edge-aware fusion does not explicitly enforce pixel-level semantic guidance at the boundary, occasionally resulting in ambiguous transition textures. Third, our framework has limited control over cast shadows, which can lead to physically inconsistent shading in certain cases.

%% file: sec/7_conclusion.tex
\section{Conclusion}
In this work, we address the challenge of satellite image composition, a task crucial for remote-sensing simulation and data augmentation. We propose a training-free diffusion-based harmonization framework \textbf{HarmoniDiff-RS}, which achieves state-of-the-art harmonization score, outperforming traditional blending and modern diffusion baselines. Importantly, the approach preserves semantic fidelity while adapting appearance and context to the target domain. Future work includes improving physical realism such as shadow consistency, leveraging higher-fidelity latent models, and extending the framework to multi-object and generative scene planning scenarios.